# Improving Clinical NLP Performance through Language Model-Generated Synthetic Clinical Data

Shan Chen, MS[1,2,3] Jack Gallifant, MBBS[4,5] Marco Guevara, MS[1,2] Yanjun Gao, PhD[6]

Majid Afshar MD MSCR[6] Timothy Miller, PhD[3] Dmitriy Dligach, PhD[7]

Danielle S. Bitterman, MD[1,2,3]

1. Artificial Intelligence in Medicine (AIM) Program, Mass General Brigham, Harvard Medical School, Boston, MA, USA
2. Department of Radiation Oncology, Brigham and Women's Hospital/Dana-Farber Cancer Institute, Boston, MA, USA
3. Computational Health Informatics Program, Boston Children's Hospital, Harvard Medical School, Boston, MA, USA
4. Institute for Medical Engineering and Science, Massachusetts Institute of Technology, Cambridge, MA, USA
5. Department of Critical Care, Guy's & St Thomas' NHS Trust, London, United Kingdom,
6. Department of Medicine, University of Wisconsin School of Medicine and Public Health, Madison, WI, USA
7. Department of Computer Science, Loyola University Chicago, Chicago, IL, USA

Brief communication 70 words: Generative models have been showing potential for producing data in mass. This study explores the enhancement of clinical natural language processing performance by utilizing synthetic data generated from advanced language models. Promising results show feasible applications in such a high-stakes domain.

Corresponding author:
Dr. Danielle S. Bitterman
Department of Radiation Oncology
Dana-Farber Cancer Institute/Brigham and Women's Hospital
75 Francis Street, Boston, MA 02115
Email: dbitterman@bwh.harvard.edu
Phone: (857) 215-1489
Fax: (617) 975-0985

Prior presentations: AMIA 2024 Informatics Summit Oral Abstract

*The authors acknowledge financial support from the Woods Foundation (DB) NIH (NIH-USA U54CA274516-01A1 (SC, DB), NIH-NIDA R01DA051464 (MA), NIH-USA R01LM012973 (TM,MA), NIH-USA R01MH126977 (TM), NIH-USA U54 TW012043-01 (JG), NIH-USA OT2OD032701 (JG).*

Disclosures:
*DSB: Associate Editor of Radiation Oncology, HemOnc.org (no financial compensation, unrelated to this work); Funding from American Association for Cancer Research (unrelated to this work). MercurialAI (scientific advisory board).*

### Introduction:

A common challenge for the development of clinical natural language processing (NLP) methods is the availability of large annotated datasets for model training, fine-tuning, and evaluation. Traditional annotation processes are time-consuming, expensive, and often require expert medical knowledge, creating significant research and benchmark development constraints.[1] Furthermore, concerns around patient privacy and data governance further complicate the sharing of large clinical datasets and limit the development of generalizable models.[2,3]

There has been increasing interest in synthetic-based approaches to overcome these constraints, with generative adversarial network-based methods having already shown promise in EHR-based tabular data.[4] However, traditional text-based synthetic data methods like paraphrasing and word swapping have been constrained by their limited semantic and style variety and challenges surrounding grammatical errors.[5–8] The advances in large language models (LLMs), which excel at following natural language instructions to generate fluent text, not only offer promise in solving clinical tasks [9–12], but may also be well-suited to synthetic text generation that is high enough quality to augment manually labeled datasets for NLP model development. [5–8] Such synthetic datasets could accelerate the development of high-performing clinical NLP models by minimizing clinical data and human effort requirements.

This study proposes a novel method for generating synthetic annotated clinical text datasets using LLMs, and evaluates their impact on downstream task performance compared to those trained on gold-standard, expert-annotated datasets. We introduce a novel approach of label correction, which is an active learning step that we apply to enhance synthetic dataset quality. We demonstrate the efficacy of our synthetic data augmentation step on both NLP benchmarks and real-world long document clinical datasets.

### Methods:

Our research evaluated synthetic data's value for existing curated clinical benchmark tasks (Figure 1a) and for a real-world, long-document clinical task. For the curated tasks, we used the following three clinical NLP tasks from DR.BENCH[13], which was developed using the MIMIC III dataset[14]: medical natural language inference (MedNLI), Assessment and Plan relation labeling (A/P Reasoning), and problem list summarization (ProbSum). We employed two versions of the Llama-2 LLM for synthetic data generation: one with 7 billion parameters and another with 70 billion parameters[15].

For synthetic data generation, a fixed subset of the training set served as exemplars for LLM-generated prompts to generate new data. The gold standard MedNLI dataset consists of 11,232 annotated data points, of which 20% were used as exemplars. The A/P Reasoning dataset consists of 4,633 annotated data points, of which 100% were used as exemplars. The ProbSumm dataset consists of 600 data points, of which 50% were used as exemplars.

For the label correction step, these exemplars were also used to fine-tune models for the DR.BENCH tasks, referred to as "label corrector" models, to improve the quality of synthetic data labels. We generated an amount of synthetic data equal to the size of the original training set for each task. The amount of data used to generate synthetic data for each dataset was chosen by the amount of data needed to obtain a stable label corrector. [16]

For the two classification tasks (MedNLI & A/P Reasoning), we fine-tuned the FLan-T5-3B[17] model with low-rank adaptors (LoRA)[18]. For the text generation task (ProbSumm), we fine-tuned the Llama-2-7B model. We evaluated two approaches to using synthetic data: replacing the gold-standard training dataset entirely with the synthetic dataset and augmenting the gold-standard training dataset with the synthetic dataset. Performance was assessed on the held-out test sets for each task.

For the real-world, long-document clinical task (Figure 1b), we focused on a practical clinical task involving detecting esophagitis and its severity in cancer patient notes.[19] Adopting a similar approach, we fine-tuned PubMedBERT [20] for three classification tasks related to this clinical condition. A subset of 200 out of 1,243 gold-labeled notes from the original training set was utilized for synthetic data generation only. We used a HIPAA-compliant GPT-3.5 Turbo 0613 model, accessed through the MGB-Azure OpenAI service, for summarization (to overcome long document challenges) and synthetic data generation via in-context learning, ensuring the study's adherence to privacy regulations.

We used T5/Llama models as classifiers for the DR.BENCH series of tasks and used PubMedBERT for our esophagitis grading tasks simply because the previous state-of-the-art for these tasks deployed the same models for better direct comparisons. We used GPT-3.5 to generate synthetic data for the esophagitis grading tasks because Llama-2 failed to generate reliable summaries for this real-world task.

This study was approved by the Mass General Brigham IRB.
You can find all our detailed prompts for the synthetic data generation step at https://github.com/AIM-Harvard/fake2real. Due to real PHI limitations, we cannot share our generated synthetic data.

**Results:**
The performance of clinical models was evaluated in three scenarios: (1) fine-tuning with only gold-standard data (i.e., expert-annotated real clinical text); (2) fine-tuning with only synthetic data; (3) synthetic data generation with few-shot examples and no label corrections; and (4) fine-tuning with an augmented dataset consisting of gold plus synthetic data. Performance was benchmarked using the Diagnostic Reasoning Benchmark (DR.BENCH[13]) dataset, which includes tasks such as medical natural language inference (MedNLI), Assessment and Plan relation labeling (A/P Reasoning), and problem list summarization (ProbSum); see Table 1.1.

Among all clinical benchmark tasks, we observed a large drop in performance when using synthetic data alone without label correction. However, with label correction, incorporating

### Table 1.1 Clinical benchmark datasets

| Approach | The model used to generate raw synthetic data | Label corrector model used | Gold data used for training (%)* | ProbSumm: Rouge-L | A/P Reasoning: Macro F1 | MedNLI: Macro F1 |
|---|---|---|---|---|---|---|
| Gold Only | NA - Previous SOTA | NA | 100 | 28.55 | 0.77 | 0.85 |
| | NA - 100% human-annotated data | NA | 100 | **28.67** | 0.79 | **0.86** |
| | NA - n% human-annotated data | NA | 50/100/20 | 24.55 | 0.79 | 0.85 |
| Synthetic Only (w/o Label Correction) | Llama 2 - 7b | None | NA | 6.72 | 0.57 | 0.51 |
| | Llama 2 - 70b | None | NA | 9.52 | 0.63 | 0.60 |
| Synthetic Only (w/ Label Correction) | Llama 2 - 7b | Yes | NA | 27.88 | 0.71 | 0.83 |
| | Llama 2 - 70b | Yes | NA | 27.73 | 0.73 | 0.84 |
| Gold + Synthetic | Llama 2 - 7b | Yes | 50/100/20 | 26.61 | **0.80** | 0.84 |
| | Llama 2 - 70b | Yes | 50/100/20 | 26.77 | 0.80 | 0.84 |

### Table 1.2 Esophagitis

| Approach | The model used to generate raw synthetic data | Label corrector model used | Gold data used for training (n) | Task 1** Macro F1 | Task 2** Macro F1 | Task 3** Macro F1 |
|---|---|---|---|---|---|---|
| Real data Only | NA - Previous SOTA | NA | 1243+2420*** | 0.92 | **0.82** | 0.74 |
| | NA - 100% human-annotated data | NA | 1243 | 0.88 | 0.79 | 0.66 |
| | NA - n% human-annotated data | NA | 200 | 0.71 | 0.55 | 0.57 |
| Summarized Gold | NA | NA | 1243 | 0.82 | 0.77 | 0.75 |
| Synthetic Only (w/o Label Correction) | GPT3.5-turbo | NA | NA | 0.83 | 0.74 | 0.67 |
| Gold + Synthetic | GPT3.5-turbo | Yes | 1243 | **0.92** | 0.80 | **0.76** |

NA: not applicable; SOTA: state of the art
*The same exact data was used to generate exemplars and to train the final models (order for table 1.1 as ProbSumm, A/P reasoning, and MedNLI)
** Task 1: Presence or absence of esophagitis. Task 2: Presence or absence of severe esophagitis. Task 3: Presence of no esophagitis vs. mild esophagitis vs. severe esophagitis.
*** For the previous SOTA on this task, we used 1243 gold-labeled notes augmented with 2420 silver-labeled real clinical notes to train the final model

synthetic data via augmentation and replacement demonstrated competitive results across all tasks. The overall trends in performance differences between the approaches were similar across the three tasks, regardless of the model sizes and types used to generate the synthetic data. Notably, for the A/P Reasoning task, the augmentation approach with synthetic data exceeded the prior state-of-the-art performance achieved with gold-only data. In the case of the ProbSumm and MedNLI tasks, synthetic-only datasets, when used with label correction, approached the performance of the gold-only dataset.

The generalisability of these results to real-world clinical tasks was also evaluated using a disease classification task; more specifically, grading esophagitis severity in cancer patient notes (Table 1.2). Using over 3600 annotated real patient notes, the macro F1 scores for the three classification tasks were 0.92, 0.82, and 0.74, respectively; observing a small performance drop using a gold-only summary compared to the gold-only model across the three tasks. In contrast, the model fine-tuned on only synthetic data (labeled using the label corrected created with 200 gold-labeled notes) outperformed the model fine-tuned on the same 200 gold-labeled notes only, and reached comparable performance to the model fine-tuned with all 1243 gold-labeled notes. Furthermore, the synthetic-only model outperformed in-context learning approaches using GPT-3.5. The highest performance was observed using the augmentation strategy combining gold and synthetic data, which realized scores comparable to those of using data augmentation with real silver-labeled clinic notes.

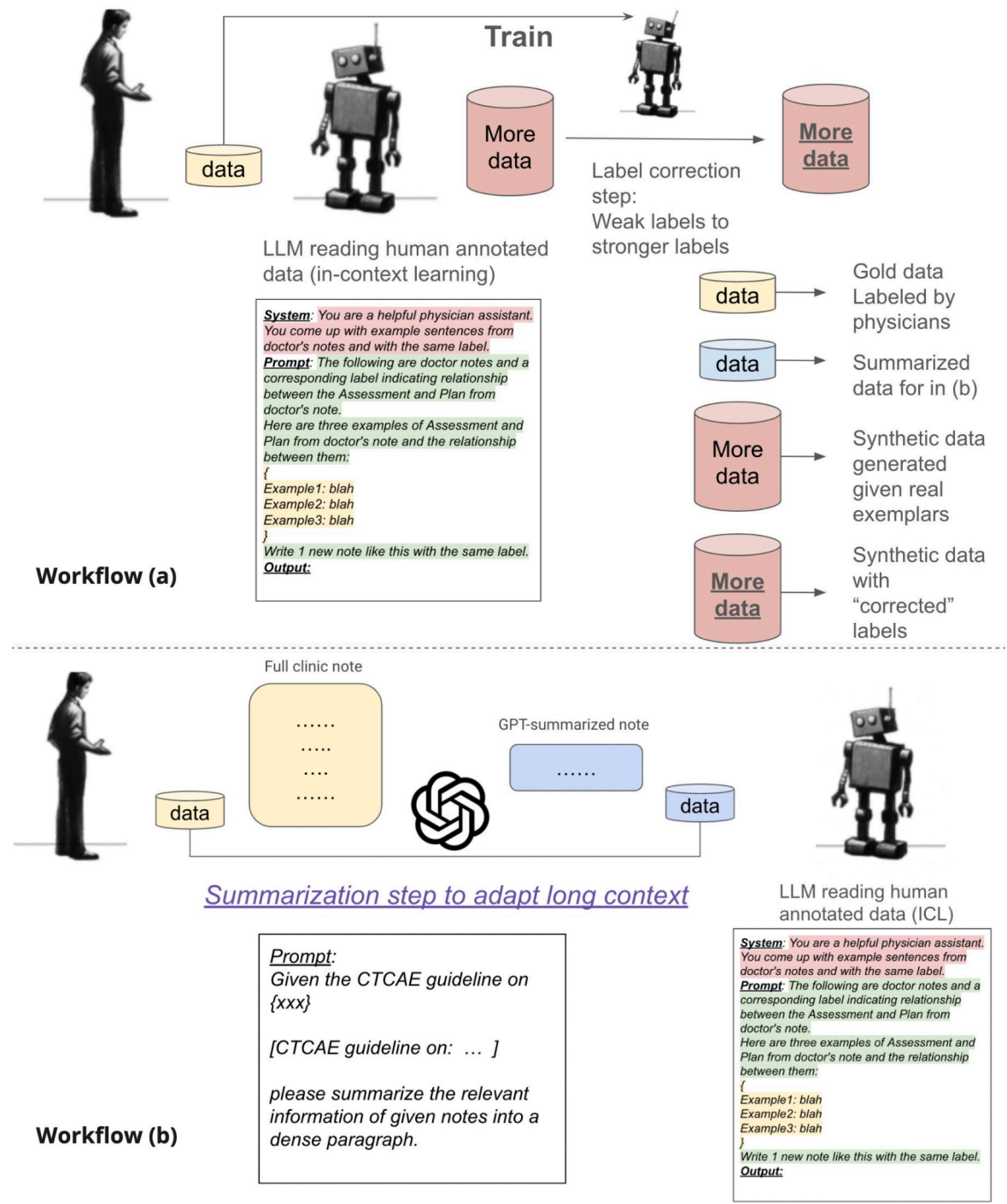

**Figure 1.** a) Our proposed workflow for synthetic data generation with label correction. b) Note-level esophagitis classification included an additional summarization pre-processing step, where GPT-3.5-turbo was used to first summarize relevant information followed by synthetic summary generation. *Illustration style inspired by OpenAI Weak2strong project.*

**Discussion:**
We show that LLMs can generate synthetic clinical datasets that approach or exceed state-of-the-art performance on benchmarks and real-world tasks, particularly when used to augment expert-labeled examples. This approach could mitigate the challenges of generating large annotated clinical NLP datasets, which have applications across biomedical research and clinical care. Our novel label corrector approach yielded substantial performance improvements compared to synthetic texts created using in-context learning alone.

One of the key implications of our work is the potential for reducing reliance on large volumes of real clinical data, which is often difficult to obtain due to privacy concerns and the need for expert annotation. By generating synthetic data that closely mimics real clinical text, our method offers a scalable solution to these challenges while reducing annotation requirements; exemplified by achieving comparable performance with less than one-sixth of the annotated data in our real-world clinical task. Further, as AI-generated clinical text becomes more prevalent with LLM-enabled documentation, synthetic-based approaches could play an important role in fine-tuning models and generating clinical-based benchmarking methods to assure their robustness and safety in real-world settings.

**Conclusion:**
Overall, our study underscores the significant potential of synthetic data generated by LLMs to enhance the performance of downstream clinical NLP tasks. By incorporating synthetic data alongside expert-annotated datasets, our methods could help address critical challenges such as data scarcity and the intensive demand for expert annotation.

Our findings highlight the importance of continuous refinement in synthetic data generation and label correction techniques. This research demonstrates the promise of LLM-enabled data augmentation and model training in clinical NLP, and opens avenues for further investigation into the refinement and integration of synthetic data. Future research directions include exploring synthetic proxy data-sharing across institutions, multi-institutional synthetic benchmarking, and assessing potential biases introduced by synthetic data.

**Role of the funding source:** The study's funders had no role in the design, data collection, analysis, interpretation, or report writing.